\documentclass[letterpaper]{article} 
\usepackage{aaai2026}  
\usepackage{times}  
\usepackage{helvet}  
\usepackage{courier}  
\usepackage[hyphens]{url}  
\usepackage{graphicx} 
\usepackage{natbib}  
\usepackage{caption} 
\usepackage{algorithm}
\usepackage{algorithmic}
\usepackage{xcolor}

\usepackage{newfloat}
\usepackage{listings}
\usepackage{graphicx}
\usepackage{multirow}
\usepackage{subfigure}
\usepackage{booktabs}
\usepackage{amsmath}
\DeclareCaptionStyle{ruled}{labelfont=normalfont,labelsep=colon,strut=off} 
\floatstyle{ruled}
\newfloat{listing}{tb}{lst}{}
\floatname{listing}{Listing}
\title{BSIPipe: Bidirectional Sequence-Level Interleaved Pipeline Parallelism for Long-Sequence Large Language Model Training}

\author{
	Yuzhou Ni\textsuperscript{\rm 1,\rm 2}, Ling Chen\textsuperscript{\rm 1,\rm 2,\thanks{Corresponding Author}}\\
}
\affiliations{
	\textsuperscript{\rm 1}State Key Laboratory of Blockchain and Data Security, Zhejiang University\\
	\textsuperscript{\rm 2}College of Computer Science and Technology, Zhejiang University
	
	\{yuzhouni, lingchen\}@cs.zju.edu.cn
}

\usepackage{bibentry}

\begin{document}

\maketitle

\begin{abstract}
Pipeline parallelism has become a critical technique for 
training large language models. To address the throughput 
limitation of existing sequence-level methods that schedule 
subsequences unidirectionally, we propose bidirectional 
sequence-level interleaved pipeline parallelism (BSIPipe), 
which co-schedules two opposing V-shaped interleaved 
pipelines at the subsequence level. To bound the activation 
accumulation from subsequence scaling, BSIPipe introduces a 
basic-unit concatenation mechanism that keeps activation 
memory invariant to the subsequence count. To eliminate 
communication deadlock risk from bidirectional scheduling, 
BSIPipe employs a communication optimization strategy that 
combines topology-aware device mapping with 
communication-computation overlap. In addition, BSIPipe composes with tensor parallelism to 
overcome the memory limitation of single-dimensional 
parallelism, neutralizing its bidirectional memory overhead 
while maintaining throughput. Experiments on GPT and LLaMA3 
demonstrate that BSIPipe achieves average speedups of 
1.11$\times$ to 1.13$\times$ in long-sequence training over 
SOTA sequence-level baselines.
\end{abstract}


\section{Introduction}

The rapid advancement of Transformer-based large 
language models (LLMs), e.g., GPT and 
LLaMA~\cite{touvron2023llama}, has driven remarkable breakthroughs 
across various domains, including natural language processing~\cite{instructgpt}, code 
generation~\cite{codellama}, and intelligent question answering~\cite{chowdhury2025handling}. As model scale grows, training these 
models on a single device has become practically infeasible: 
GPT-3~\cite{gpt3}, for instance, requires over 350\,GB of memory at half 
precision and would demand approximately 288 years to train on a single 
NVIDIA V100 GPU~\cite{interleaved}. Beyond model scale, 
the growing need to capture long-range dependencies 
is driving training sequence lengths substantially higher, with 
state-of-the-art (SOTA) LLMs trained on sequences ranging from thousands to tens 
of thousands of tokens~\cite{chen2023longlora,liu2024ringattention}.
The growth along these two dimensions jointly imposes severe demands on memory and 
computational throughput, making efficient distributed training indispensable for modern LLM development.

As a key parallelism paradigm in distributed training, pipeline 
parallelism (PP)~\cite{guan2024advances} partitions a model into a sequence of 
stages across multiple devices. Each batch is then split 
into multiple micro-batches that traverse these stages in a pipelined 
manner. This design alleviates the single-device memory bottleneck while 
overlapping communication with computation, and has therefore been widely 
adopted in LLM training~\cite{deepseekv3,hui2024qwen2}. As sequence lengths grow, the computation time per micro-batch and the 
device idle time (also called pipeline bubbles) increase correspondingly, 
hindering further throughput gains. To 
mitigate this, sequence-level pipeline parallelism~\cite{terapipe} further 
partitions each micro-batch into multiple subsequences, which serve as 
finer-grained schedulable units. This partitioning increases the number 
of subsequences traversing the pipeline, thereby reducing the bubble 
ratio. 
Recent studies further explore the 1F1B schedule and its interleaved 
variant to improve training throughput~\cite{seq1f1b}; others investigate 
memory optimization through bubble insertion or recomputation with 
activation offloading~\cite{mepipe,slimpipe}.
However, all these 1F1B-based methods introduce substantial pipeline bubbles during the warmup 
and cooldown phases due to strict inter-stage dependencies and periodic synchronization, 
reducing device utilization and limiting training efficiency.

To address this limitation, a natural approach is to introduce 
bidirectional scheduling, which co-schedules two opposing pipelines to 
better utilize idle devices. Despite its potential, integrating 
bidirectional scheduling with fine-grained sequence-level scheduling is non-trivial,
posing three key challenges:
\textbf{(1) Activation accumulation from subsequence scaling.} Under 
unidirectional 1F1B scheduling, forward and backward computations 
alternate, and activations can be released once their backward 
passes complete, so increasing the number of subsequences does not raise 
memory consumption. In contrast, bidirectional scheduling exhibits a 
forward-then-backward execution boundary: as the number of subsequences 
grows, forward activations persistently accumulate in memory until the 
backward phase begins, causing memory consumption to grow linearly with 
the number of subsequences. 
\textbf{(2) Communication deadlock risk from bidirectional scheduling.} 
Under unidirectional 1F1B scheduling, inter-device communication has a 
fixed direction, and adjacent devices consistently maintain a 
well-ordered ``one-send-one-receive'' pattern. By contrast, 
bidirectional scheduling concurrently drives two 
pipelines in opposite directions, in which adjacent devices 
may simultaneously issue send operations and thus fail to 
receive, triggering communication deadlocks. 
\textbf{(3) Memory limitation of single-dimensional parallelism.} 
A single parallelism dimension struggles to balance computational 
efficiency and memory utilization, and bidirectional scheduling 
further amplifies memory pressure while boosting training 
throughput. Consequently, memory limitation remains a critical 
bottleneck for practical deployment.

To this end, we propose \textbf{BSIPipe}, a \textbf{B}idirectional  \textbf{S}equence-Level \textbf{I}nterleaved 
\textbf{P}ipeline Parallelism. To the best of our knowledge, 
BSIPipe is the first work to introduce bidirectional scheduling into 
sequence-level pipeline parallelism. By co-scheduling two 
V-shaped interleaved pipelines running in opposite directions, BSIPipe 
achieves almost zero-bubble execution in the steady phase and substantially 
mitigates pipeline bubble overhead in the warmup and cooldown phases. The 
main contributions are as follows:

\begin{itemize}
  \item We propose a basic-unit concatenation mechanism that takes the 
  schedule satisfying the minimum number of subsequences required for 
  the warmup phase as a basic unit, which extends along the subsequence 
  dimension by concatenating multiple basic units. By bounding activation 
  accumulation within a single basic unit, we alleviate memory pressure 
  caused by the forward-then-backward execution pattern.

  \item We propose a communication optimization strategy for bidirectional 
  scheduling, which combines an early-receive 
  mechanism to avoid deadlocks, topology-aware device mapping to confine 
  gradient synchronization within intra-node high-bandwidth interconnects, 
  and communication-computation overlap  
  to improve training throughput.

  \item We introduce tensor parallelism to construct a multi-dimensional 
    parallelism strategy, which reduces per-device parameter and activation memory as the tensor 
parallelism degree increases. In addition, we 
    further confine both gradient synchronization and tensor-parallel 
    collective communication to intra-node high-bandwidth interconnects, 
    reducing communication latency in multi-node training scenarios.

  \item Extensive experiments show that BSIPipe achieves the highest 
  training throughput on GPT and LLaMA3 in long-sequence scenarios, 
  delivering average speedups of 1.11${\times}$--1.13${\times}$ over 
  SOTA sequence-level methods.
\end{itemize}

\section{Related Work}
\begin{figure*}[t]
  \centering
  \includegraphics[width=\textwidth]{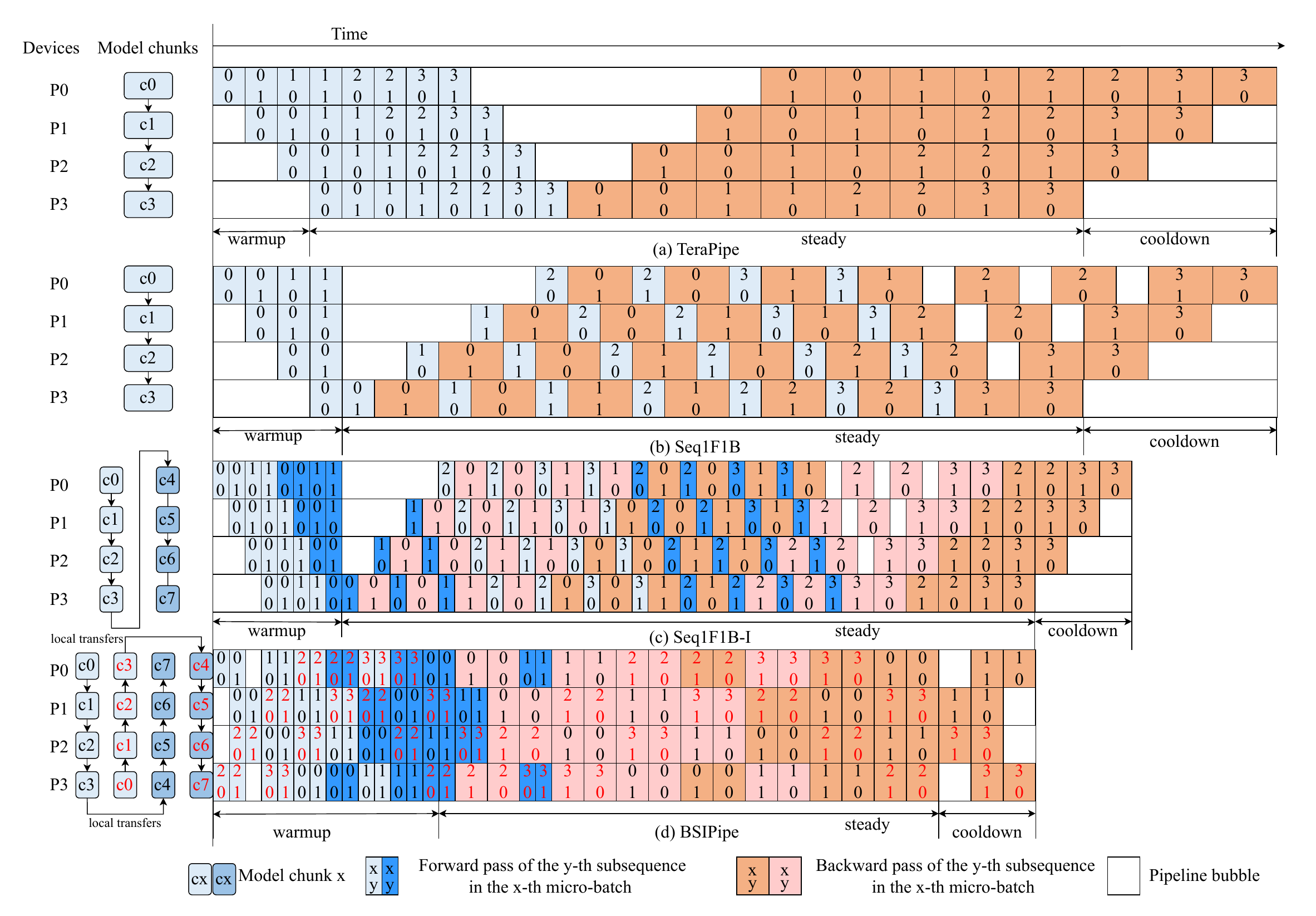}
\caption{Comparison of representative sequence-level pipeline parallelism 
methods. (a) TeraPipe schedules subsequences in a forward-then-backward 
manner. (b) Seq1F1B applies the 1F1B schedule at the subsequence level. 
(c) Seq1F1B-I further combines this with model chunk interleaving, where 
shades of each color denote computations of different model chunks on the 
same device. (d) BSIPipe (ours) co-schedules two opposing V-shaped 
interleaved pipelines at the subsequence level. }
\label{fig:bsipipe_compare}
\end{figure*}
\subsection{Tensor Parallelism}
Tensor parallelism (TP) partitions compute-intensive operators 
across devices and synchronizes results via collective 
communication. General-purpose sharding frameworks~\cite{gshard} 
support flexible partitioning schemes, while Megatron-LM~\cite{interleaved} 
establishes the canonical 1D TP scheme for Transformer layers. 
To reduce activation memory, Megatron-SP~\cite{sp} partitions 
inter-layer activations along the sequence dimension, while 
context parallelism~\cite{cp}, Ring 
Attention~\cite{liu2024ringattention}, and 
DeepSpeed-Ulysses~\cite{ulysses} further partition intra-layer 
attention activations for long-context training. 
2D TP schemes such as Optimus~\cite{optimus} adapt classical 
parallel matrix-multiplication algorithms to 2D device 
layouts, reducing memory overhead over the 
1D partitioning of Megatron-LM. Overlap techniques such as 
Flux~\cite{flux} fuse tile-level collectives into 
matrix-multiplication kernels to hide their latency. 
These advances establish TP as a mature and orthogonal 
parallelism dimension, complementary to PP 
for LLM training.

\subsection{Pipeline Parallelism}
Pipeline parallelism is broadly divided into synchronous and 
asynchronous paradigms. Synchronous methods dominate LLM training, which flush the 
pipeline each iteration to preserve strict optimizer semantics.
GPipe~\cite{gpipe} introduces micro-batch 
partitioning but lets activations accumulate until the backward 
phase; DAPPLE~\cite{dapple} introduces the 1F1B schedule that 
releases activations promptly and has since become the mainstream 
foundation; 1F1B-I~\cite{interleaved} splits each stage 
into smaller chunks distributed in an interleaved layout to 
further reduce the bubble ratio. These methods schedule 
micro-batches unidirectionally, leaving downstream devices idle 
during warmup and cooldown. To overcome this limitation, 
ZeroBubble~\cite{zerobubble} decomposes the backward pass into 
input-gradient and weight-gradient computations to fill 
remaining bubbles, while Chimera~\cite{chimera} and its 
V-shaped extension BitPipe~\cite{bitpipe} co-schedule two 
opposing pipelines to occupy these idle devices, at the cost 
of doubled parameter memory. All of these 
methods operate at the micro-batch level: as sequence length 
grows, pipeline bubbles grow correspondingly, motivating 
finer-grained scheduling along the sequence dimension.
Asynchronous methods such as 
PipeDream~\cite{pipedream}, PipeDream-2BW~\cite{pipedream-2bw}, 
and PipeMare~\cite{pipemare} eliminate pipeline flushes for 
bubble-free execution at the cost of relaxed convergence 
guarantees, which lie outside the scope of this work. 

\subsection{Sequence-Level Pipeline Parallelism}

Sequence-level pipeline parallelism partitions each micro-batch 
into multiple subsequences as 
finer-grained schedulable units, reducing the bubble ratio in 
long-sequence training. As illustrated in 
Figure~\ref{fig:bsipipe_compare}, TeraPipe~\cite{terapipe} 
pioneers this approach by extending the forward-then-backward 
schedule to the subsequence level, while Seq1F1B and its 
interleaved variant~\cite{seq1f1b} adopt the 1F1B schedule and 
model chunk interleaving to release activations promptly and 
reduce bubbles in the warmup and cooldown phases. Building on 
these, MePipe~\cite{mepipe} actively inserts bubbles to reduce 
peak activation memory at the cost of throughput, and 
SlimPipe~\cite{slimpipe} combines recomputation with activation 
offloading to further reduce the memory footprint. These methods 
all schedule subsequences unidirectionally, leaving downstream 
devices substantially idle during warmup and cooldown.

\begin{table}[t]
  \centering
  \footnotesize
    \setlength{\tabcolsep}{2pt}

  \begin{tabular}{lccc}
    \toprule
    Method & Bubble Ratio & Activation & Parameter \\
    \midrule
    TeraPipe & $\frac{w-1}{N_s+w-1}$ & $N_s M_a$ & $M_\Theta$ \\
    Seq1F1B & $\frac{w-1}{N_s+w-1}$ & $[M_a, w M_a]$ & $M_\Theta$ \\
    Seq1F1B-I & $\frac{w-1}{2N_s+w-1}$ & $[\tfrac{w+1}{2} M_a, w M_a]$ & $M_\Theta$ \\
    \textbf{BSIPipe (Ours)} & $\frac{w-1}{3N_s+w-1}$ & $[(w+3) M_a, 2w M_a]$ & $2 M_\Theta$ \\
    \bottomrule
  \end{tabular}
  \caption{Comparison of bubble ratio and per-device memory across 
  sequence-level pipeline parallelism methods.}
  \label{tab:bubble_memory}
\end{table}

Motivated by the capability of bidirectional scheduling to better utilize idle 
devices, in this paper, we propose 
BSIPipe, the first method to integrate bidirectional scheduling 
with sequence-level execution. Specifically, a basic-unit 
concatenation mechanism is designed to bound per-device 
activation memory regardless of the subsequence count. Then, 
a communication optimization strategy is designed to 
eliminate cross-pipeline communication deadlocks and 
maximize communication-computation overlap.  
Finally, a multi-dimensional parallelism strategy is 
designed to integrate BSIPipe with TP for 
further memory reduction.

\section{Methodology}
\subsection{Bidirectional V-Shaped Subsequence Scheduling}
\label{sec:bsipipe_schedule}

BSIPipe co-schedules two opposing V-shaped pipelines, mirrored 
along the device axis, to occupy downstream devices that would 
otherwise remain idle under unidirectional scheduling. As illustrated 
in Figure~\ref{fig:bsipipe_compare}(d), the two pipelines launch 
subsequences from opposite ends of the device axis during warmup, run concurrently 
through the steady phase to achieve almost zero-bubble execution, 
and complete their cooldown in parallel from both ends. Within 
each pipeline, the majority of forward passes are executed before backward passes begin, forming a forward-then-backward execution 
boundary. After all forward and backward passes complete, devices 
holding the same model chunk perform gradient synchronization, 
followed by a parameter update.

\textbf{V-shaped chunk layout.} Let $w$ denote the pipeline depth and $p$ 
the interleaving factor (the number of model chunks per device). The 
model is partitioned into $pw$ chunks $c_0, \ldots, c_{pw-1}$. For the 
standard configuration $p=2$ illustrated in 
Figure~\ref{fig:bsipipe_compare}(d), the down pipeline assigns chunks 
$c_0, \ldots, c_{w-1}$ to devices $0, \ldots, w-1$ in order and chunks 
$c_w, \ldots, c_{2w-1}$ to devices $w-1, \ldots, 0$ in reverse, forming a 
V-shaped layout; the up pipeline mirrors this layout. Unlike the 
sequential chunk assignment in Seq1F1B-I, the V-shape places one pair of 
adjacent chunks ($c_{w-1}$ and $c_w$) within the same device at 
the V's turning point, reducing the corresponding inter-device transfer 
to an intra-device data copy.

\textbf{Subsequence partitioning.} BSIPipe partitions each 
micro-batch into $k$ compute-balanced subsequences following 
Seq1F1B, so that all subsequences yield equal FLOPs and the 
inter-subsequence bubbles induced by the causal attention pattern 
are eliminated. To achieve the minimum bubble ratio, the total 
number of subsequences per iteration $N_s$ must be a multiple of 
$kw$, and the number of model layers $L$ must be a multiple of 
$pw$.

\textbf{Comparison with existing methods.} 
Table~\ref{tab:bubble_memory} compares BSIPipe against prior 
sequence-level methods in terms of bubble ratio 
and per-device memory, where $M_a$ denotes per-subsequence activation 
memory and $M_\Theta$ denotes per-stage parameter memory. BSIPipe's 
bidirectional V-shaped schedule reduces the bubble ratio to 
$\frac{w-1}{3N_s + w - 1}$. This is lower than TeraPipe and 
Seq1F1B's $\frac{w-1}{N_s + w - 1}$, and lower than Seq1F1B-I's 
$\frac{w-1}{2N_s + w - 1}$. As $w$ grows, all methods' bubble ratios increase, 
but BSIPipe is 
the least sensitive, as $N_s$ carries the largest coefficient (3) 
in BSIPipe's denominator.
Conversely, as $N_s$ grows, BSIPipe's bubble ratio decreases to 
zero most rapidly. This advantage comes at the cost of doubled 
parameter memory ($2M_\Theta$) and accumulated activation memory 
from co-scheduling two pipelines.

\subsection{Basic-Unit Concatenation}
\label{sec:concat}

\begin{figure}[tb]
  \centering
  \subfigure[Naive scaling: injecting most $N_s$ subsequences during 
  warmup causes activations to accumulate until backward begins.]{
    \includegraphics[width=0.85\linewidth]{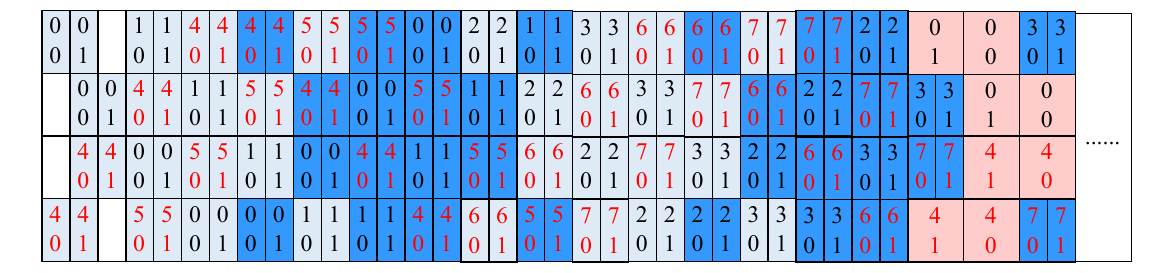}
    \label{fig:concat_naive}
  } \\
  \subfigure[Basic-unit concatenation: the warmup of each subsequent 
  unit fills the cooldown bubbles of the preceding unit.]{
    \includegraphics[width=\linewidth]{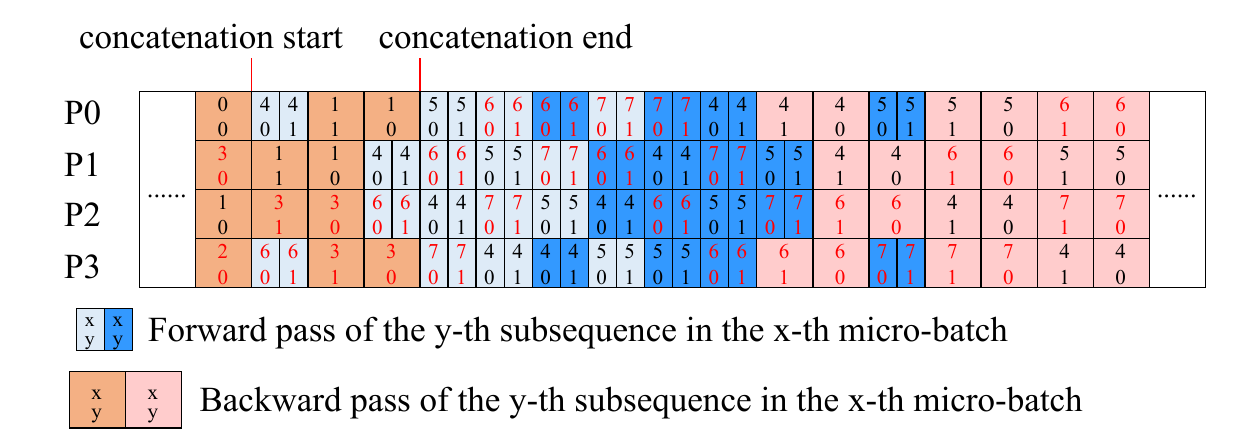}
    \label{fig:concat_unit}
  }
  \caption{Scaling BSIPipe along the subsequence dimension.}
  \label{fig:concat}
\end{figure}

Since BSIPipe's bubble ratio decreases as $N_s$ grows, scaling 
the subsequence count is an attractive approach for further 
throughput improvements. This scaling, however, is constrained by 
BSIPipe's forward-then-backward execution boundary. As illustrated in 
Figure~\ref{fig:concat_naive}, injecting all $N_s$ subsequences 
during warmup causes forward activations to accumulate in device 
memory until the backward phase begins, so per-device activation 
memory grows linearly with $N_s$ and quickly exhausts GPU memory.

To eliminate this linear growth without sacrificing scalability, 
BSIPipe concatenates multiple \emph{basic units}.  
As illustrated in Figure~\ref{fig:bsipipe_compare}(d), a basic unit is the 
smallest BSIPipe schedule that fully utilizes all devices and 
contains $kw$ subsequences. To support a target subsequence count 
$N_s$ (a multiple of $kw$), BSIPipe concatenates $N_s/kw$ basic 
units along the time axis. As illustrated in 
Figure~\ref{fig:concat_unit}, the warmup of each subsequent basic 
unit overlaps with the cooldown of the preceding unit, filling the cooldown bubbles 
with forward computation. This concatenation 
eliminates the inter-unit bubbles and bounds each device's 
activation memory to that of a single basic unit, independent of 
$N_s$.

After concatenation, only the warmup of the first basic unit and 
the cooldown of the last basic unit incur unavoidable bubbles, 
while all intermediate units achieve almost zero-bubble execution. 
Consequently, BSIPipe is highly scalable along the subsequence 
dimension: as $N_s$ grows, the bubble ratio approaches zero while 
activation memory remains bounded, enabling substantial 
throughput gains.

\begin{figure}[tb]
  \centering
  \includegraphics[width=1.1\linewidth]{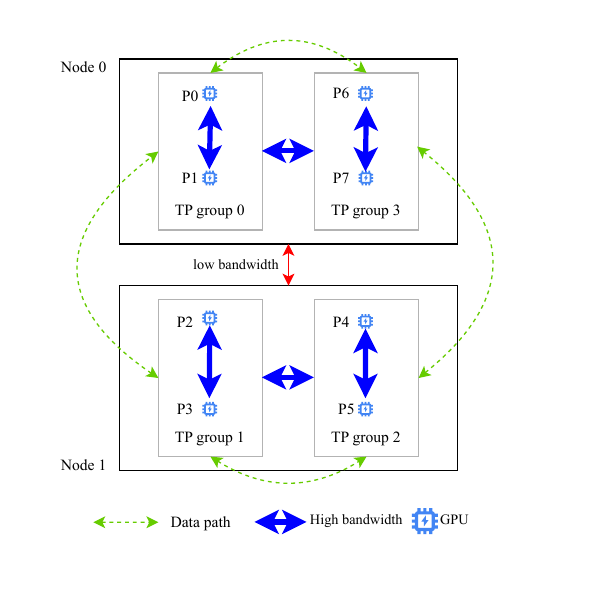}
  \vspace{-2em}
  \caption{BSIPipe device mapping under multi-dimensional parallelism. 
Each node holds two TP groups (one per pipeline) with 
mirrored-pair gradient-synchronization groups. Dashed arrows 
trace the bidirectional pipeline data path.}
  \label{fig:bsipipe_tp}
\end{figure}

\subsection{Communication Optimization}
\label{sec:comm}

In bidirectional scheduling, the two opposing pipelines run 
concurrently and each device handles communication for both. 
The resulting cross-pipeline communication can lead to 
deadlock if left unoptimized. To eliminate this risk and 
further improve communication efficiency, we apply three 
optimizations to BSIPipe.

\textbf{Topology-aware device mapping.} GPU clusters typically 
exhibit high intra-node bandwidth (NVLink, PCIe) and lower 
inter-node bandwidth~\cite{wu2026tawpipe}. The dominant 
communication cost in BSIPipe comes from parameter gradient 
synchronization between mirrored devices---those holding the same 
model chunk in the two opposing pipelines---while inter-stage 
activation transfers are much smaller in volume. BSIPipe 
co-locates each mirrored pair on the same node, confining 
gradient synchronization to intra-node
high-bandwidth interconnects, while 
only activations cross node boundaries. This substantially 
improves multi-node scalability.

\textbf{Deadlock avoidance with early receive.} The bidirectional 
schedule may cause two adjacent pipeline stages to simultaneously 
issue send operations and thus fail to receive, resulting in a 
deadlock. BSIPipe enforces complementary 
send/receive ordering between adjacent stages: one device follows 
``send-then-receive,'' and its neighbor follows 
``receive-then-send.'' This pairing guarantees that one side has 
posted a receive before the other sends, eliminating deadlock by 
construction. The ordering also enables \emph{early 
receive}---a device posts an asynchronous receive immediately 
after finishing the preceding forward pass, rather than waiting 
for the next forward pass to begin---so the activation transfer 
overlaps with subsequent computation and reduces stall time once 
the data are required.

\textbf{Forward-communication and backward-computation overlap.} Even with this 
send/receive ordering, one scenario still forces computation to wait for 
communication. Consider a device running a backward pass while 
its neighbor computes a forward activation that the device will 
need for its next forward pass. The ``send-then-receive'' 
constraint blocks the device from receiving the activation 
until the backward pass completes, delaying the next forward computation. 
To eliminate this wait, BSIPipe 
reverses the ordering whenever a backward pass overlaps with a 
neighbor's forward pass. The device executing the backward pass 
switches to ``receive-then-send,'' and its neighbor to 
``send-then-receive,'' so the forward activation arrives during 
the ongoing backward computation. This eliminates the final 
blocking wait and maximizes communication-computation overlap.

\subsection{Multi-Dimensional Parallelism}
\label{sec:multi_dim}

To further reduce per-device memory, BSIPipe composes with TP~\cite{sp} 
along an orthogonal dimension. TP 
splits each Transformer layer's parameters and activations across 
$D_t$ devices, reducing per-device parameter and activation 
memory by a factor of $D_t$.

Combining BSIPipe with TP introduces two high-frequency 
communication patterns. Within each layer, the TP group performs 
AllGather and ReduceScatter collectives~\cite{nccl}. Across the 
bidirectional pipelines, the mirrored devices synchronize gradients. 
Both patterns must remain on intra-node high-bandwidth
interconnects to avoid throughput collapse.

BSIPipe co-locates both communication patterns within the same node. 
Each node hosts two TP groups (one from each pipeline) and 
the gradient-synchronization groups for their mirrored devices. Since both types 
of communication groups remain intra-node, TP collectives and 
bidirectional gradient synchronization both run on high-bandwidth 
interconnects (e.g., NVLink). Only inter-stage activation 
transfers cross node via lower-bandwidth links (e.g., 
InfiniBand). As a result, the 
practical value of $D_t$ is bounded by the number of GPUs per 
node. Beyond that, TP collectives cross 
inter-node links and become the dominant bottleneck. As illustrated in Figure~\ref{fig:bsipipe_tp}, 
Node~0 holds the two TP groups at the V-shape's endpoints (TP 
groups 0 and 3), while Node~1 holds the two midpoint (TP groups 
1 and 2). The dashed arrows trace the bidirectional data path, 
which traverses all four TP groups in opposite directions for 
the two pipelines.

Under this combined parallelism, per-device parameter memory 
decreases from $2M_\Theta$ to $2M_\Theta/D_t$, neutralizing 
BSIPipe's bidirectional memory overhead at $D_t \geq 2$ while 
preserving its low bubble ratio.

\begin{table*}[tb]
\centering
\footnotesize
\setlength{\tabcolsep}{1.7pt}

\begin{tabular}{l cccc ccc cccc ccc}
\toprule
\multirow{3}{*}{\textbf{Method}} 
& \multicolumn{7}{c}{\textbf{GPT-1.3B}} 
& \multicolumn{7}{c}{\textbf{LLaMA3-1.3B}} \\
\cmidrule(lr){2-8} \cmidrule(lr){9-15}
& \multicolumn{4}{c}{Sequence length} 
& \multicolumn{3}{c}{GPU count} 
& \multicolumn{4}{c}{Sequence length} 
& \multicolumn{3}{c}{GPU count} \\
\cmidrule(lr){2-5} \cmidrule(lr){6-8} 
\cmidrule(lr){9-12} \cmidrule(lr){13-15}
& 5K & 10K & 15K & 20K & 8 & 16 & 32 
& 5K & 10K & 15K & 20K & 8 & 16 & 32 \\
\midrule
DAPPLE\,(PPoPP 2021)   
& 81,823 & 73,190 & 66,487 & 57,233 & 32,625 & 57,233 & 81,456 
& 91,126 & 80,097 & 69,780 & 61,438 & 35,282 & 61,438 & 97,101 \\
1F1B-I\,(SC 2021)   
& 85,826 & 77,698 & 68,855 & OOM    & OOM    & OOM    & OOM    
& 95,155 & 84,876 & 74,117 & OOM    & OOM    & OOM    & OOM    \\
Chimera\,(SC 2021)   
& \textbf{95,157} & \underline{80,222} & 68,344 & 58,928 & 31,296 & 58,928 & 83,625 
& \textbf{109,155} & \underline{91,081} & \underline{75,881} & 64,431 & 34,409 & 64,431 & \underline{108,599} \\
TeraPipe\,(ICML 2021)  
& 48,404 & 46,825 & 41,993 & 37,722 & 15,322 & 37,722 & 38,915 
& 57,606 & 52,884 & 46,404 & 40,932 & 22,743 & 40,932 & 32,386 \\
Seq1F1B\,(NAACL 2025)   
& 82,023 & 76,304 & 68,378 & 61,512 & 34,433 & 61,512 & 87,564 
& 91,731 & 83,370 & 73,826 & 65,596 & 36,994 & 65,596 & 106,606 \\
Seq1F1B-I\,(NAACL 2025) 
& 80,253 & 76,801 & \underline{69,408} & \underline{62,874} & \underline{35,965} & \underline{62,874} & \underline{88,654} 
& 86,975 & 83,223 & 74,412 & \underline{66,784} & \underline{38,629} & \underline{66,784} & 104,744 \\
\textbf{BSIPipe (ours)} 
& \underline{91,601} & \textbf{86,487} & \textbf{77,265} & \textbf{68,954} & \textbf{37,110} & \textbf{68,954} & \textbf{89,235} 
& \underline{98,022} & \textbf{93,139} & \textbf{82,496} & \textbf{73,453} & \textbf{40,177} & \textbf{73,453} & \textbf{112,232} \\
\midrule
\textbf{Speedup vs.\ 2nd} 
& 0.96$\times$ & 1.08$\times$ & 1.11$\times$ & 1.10$\times$ & 1.03$\times$ & 1.10$\times$ & 1.01$\times$ 
& 0.90$\times$ & 1.02$\times$ & 1.08$\times$ & 1.10$\times$ & 1.04$\times$ & 1.10$\times$ & 1.03$\times$ \\
\bottomrule
\end{tabular}
\caption{Training throughput (tokens/s) of BSIPipe and 
baselines. The \emph{sequence length} block is evaluated on 
16 GPUs with 64 micro-batches under varying sequence length. 
The \emph{GPU count} block is evaluated at a sequence length 
of 20K with the number of micro-batches scaled proportionally 
to the GPU count. The best results are \textbf{bolded}. The 
second-best results are \underline{underlined}.}
\label{tab-throughput}
\end{table*}

\section{Experiments}

\subsection{Experimental Setup}

\textbf{Hardware, software, and implementation.} We conduct all 
experiments on a cluster of four servers (8 NVIDIA A800 GPUs 
per server, NVLink at 400~GB/s intra-node, 200~Gbps InfiniBand 
inter-node). The software stack consists of PyTorch~2.3.0, 
Transformer-Engine~1.4.0, and CUDA~12.2 on Ubuntu~20.04. We 
implement BSIPipe within Megatron-LM~\cite{interleaved} and 
integrate all baselines into the same framework to ensure fair 
comparison.

\textbf{Models, dataset, and metrics.} We train GPT-1.3B and 
LLaMA3-1.3B, each comprising 64 Transformer layers, a hidden 
size of 1280, and 32 attention heads. The dataset is 
OpenWebText~\cite{openwebtext}, with sequence lengths ranging 
from 5K to 20K tokens and a micro-batch size of 1. We evaluate 
all approaches in terms of training throughput (tokens per 
second) and peak per-device memory usage (GiB). Following 
standard practice in pipeline 
parallelism~\cite{guan2024advances}, throughput serves as a 
practical proxy for the bubble ratio. Each configuration runs 
for 15 iterations, and we take the average throughput over the 
final two iterations.

\textbf{Baselines.} We compare BSIPipe against six 
representative pipeline parallelism methods:
\begin{itemize}
\item \textbf{Micro-batch level:} DAPPLE~\cite{dapple} 
(canonical 1F1B), 1F1B-I~\cite{interleaved} (interleaved 1F1B 
with smaller model chunks), and Chimera~\cite{chimera} 
(bidirectional scheduling).

\item \textbf{Sequence level:} TeraPipe~\cite{terapipe} 
(forward-then-backward at the subsequence level), 
Seq1F1B~\cite{seq1f1b} (sequence-level 1F1B), and 
Seq1F1B-I~\cite{seq1f1b} (interleaved Seq1F1B).
\end{itemize}
For each baseline, we obtain the best configuration by 
grid-searching the interleaving factor over $\{1, 2, 4\}$ and 
the subsequence count over $\{1, 2, 4\}$. In our experiments, 
interleaved methods (1F1B-I, Seq1F1B-I, and BSIPipe) achieve 
the best throughput at an interleaving factor of 2, while 
the others are fixed at 1, as they do not incorporate 
interleaving. Analogously, sequence-level methods (TeraPipe, 
Seq1F1B, Seq1F1B-I, and BSIPipe) achieve the best throughput 
at a subsequence count of 2, while the others are fixed at 
1, as they do not partition subsequences.

\begin{table}[tb]
\centering
\footnotesize
\setlength{\tabcolsep}{1.7pt}

\begin{tabular}{lcccccc}
\toprule
\multirow{2}{*}{\textbf{Model}} 
& \multicolumn{2}{c}{$\mathbf{(16, 1)}$} 
& \multicolumn{2}{c}{$\mathbf{(8, 2)}$} 
& \multicolumn{2}{c}{$\mathbf{(4, 4)}$} \\
\cmidrule(lr){2-3} \cmidrule(lr){4-5} \cmidrule(lr){6-7}
& Tput & Memory & Tput & Memory & Tput & Memory \\
\midrule
GPT    & 68,954 & 49.8--60.1 
       & \textbf{69,131} & 28.3--31.7 
       & 63,917 & \textbf{17.1--18.4} \\
LLaMA3 & 73,453 & 55.8--67.8 
       & \textbf{74,541} & 31.4--35.5 
       & 68,565 & \textbf{19.4--19.7} \\
\bottomrule
\end{tabular}
\caption{Training throughput (Tput, tokens/s) and per-device 
peak memory range (GiB) of BSIPipe under different 
$(w, D_t)$ configurations on 16 GPUs at a sequence length of 
20K. The best results are \textbf{bolded}.}
\label{tab-pp-tp}
\end{table}

\subsection{Performance Comparison}

We evaluate BSIPipe along two scaling dimensions: sequence 
length at a fixed GPU count (16 GPUs, 64 micro-batches), and 
GPU count at a fixed sequence length (20K, with the number of 
micro-batches scaled proportionally to the GPU count).

\textbf{Throughput.} The throughput comparison results are 
shown in Table~\ref{tab-throughput}, from which we can derive the following observations: (1) Against 
the strongest sequence-level baselines 
Seq1F1B and 
Seq1F1B-I, BSIPipe attains average speedups of 1.12$\times$ 
to 1.13$\times$ on GPT and 1.11$\times$ against both on 
LLaMA3, confirming the gain of bidirectional scheduling at 
the subsequence level. (2) BSIPipe consistently achieves the highest throughput at 
sequence lengths $\geq$10K, with speedups of 1.08$\times$ to 
1.11$\times$ over the second-best method on GPT and 
1.02$\times$ to 1.10$\times$ on LLaMA3. The advantage of 
BSIPipe over the second-best method generally grows with 
sequence length, as a larger number of subsequences amplifies 
the benefit of BSIPipe's lower bubble ratio. (3) At 5K, BSIPipe 
trails Chimera by a small margin, as subsequence partitioning 
at such short sequences yields small per-subsequence tensor 
shapes that reduce GPU kernel efficiency and partially offset 
the gain from a lower bubble ratio. (4) At a sequence length of 20K, BSIPipe maintains the 
highest throughput from 8 to 32 GPUs, with the advantage 
peaking on 16 GPUs. On 32 GPUs, the deeper pipeline introduces 
more cross-stage communication and narrows the speedup margin, 
yet BSIPipe remains the leader, demonstrating robust scaling 
of the bidirectional V-shaped schedule across cluster sizes.

\begin{figure*}[tb]
\centering
\includegraphics[width=\textwidth]{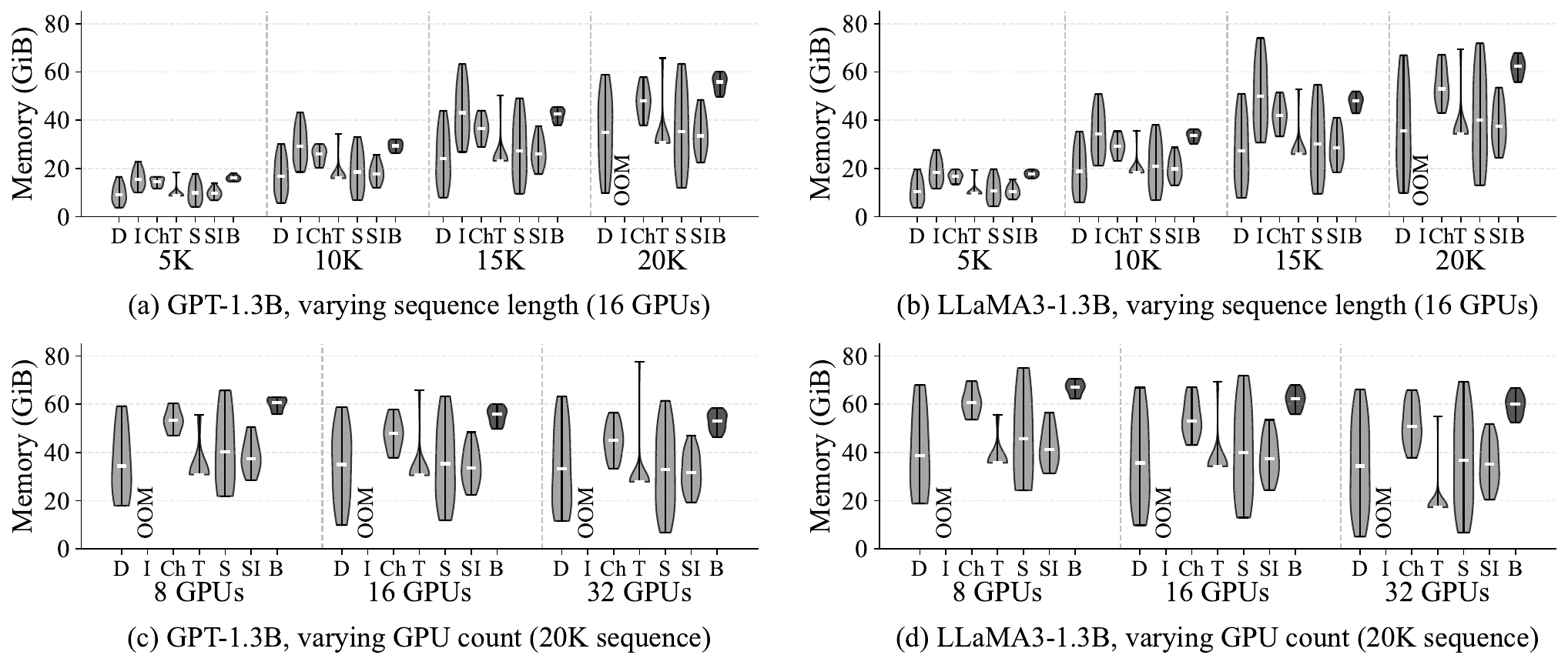}
\caption{Memory footprint distributions (GiB) of BSIPipe and 
baselines. \textbf{(a, b)} Per-device peak memory under varying 
sequence length on 16 GPUs. \textbf{(c, d)} Per-device peak 
memory under varying GPU count at a 
sequence length of 20K. Method 
abbreviations: D = DAPPLE, I = 1F1B-I, Ch = Chimera, 
T = TeraPipe, S = Seq1F1B, SI = Seq1F1B-I, and B = BSIPipe.}
\label{fig-memory}
\end{figure*}

\textbf{Memory footprint.} Figure~\ref{fig-memory} shows the memory footprint 
distributions of BSIPipe and baselines, from which we can discern the following tendencies:
(1) BSIPipe attains highly concentrated memory distributions 
across devices, varying within a tight range in every 
configuration. The reason is that the V-shaped stage assignment places 
symmetric computational loads on the down and up pipelines, 
mitigating the memory imbalance across devices that arises 
under unidirectional scheduling. 
(2) On 16 GPUs at a sequence length of 20K, BSIPipe reaches 
peak memory of 60.1 GiB on GPT and 67.9 GiB on LLaMA3, 
exceeding Seq1F1B-I by 24\% and 27\%, respectively, but 
staying below Seq1F1B and TeraPipe. The overhead over Seq1F1B-I arises from two sources. 
First, 
BSIPipe replicates the model across two pipelines for 
bidirectional execution, doubling per-device parameter memory. 
Second, even with the basic-unit concatenation mechanism, the 
forward-then-backward execution within each basic unit still 
accumulates activations during the forward phase until the 
backward phase begins. In contrast, Seq1F1B-I interleaves 
forward and backward in the steady phase and releases 
activations promptly.

\subsection{Comparison of Parallel Configurations}

To examine how BSIPipe's throughput and memory respond to 
different combinations of pipeline depth $w$ and TP degree 
$D_t$ under a fixed device budget, we run BSIPipe under three 
configurations $(w, D_t) \in \{(16, 1), (8, 2), (4, 4)\}$ on 
16 GPUs with 64 micro-batches at a sequence length of 20K.

The throughput and per-device peak memory of each 
configuration are shown in Table~\ref{tab-pp-tp}. 
The results show that: (1) The $(8, 2)$ 
configuration attains the highest throughput on both models, 
slightly outperforming $(16, 1)$. The reason is that 
$(8, 2)$ strikes a balance between pipeline depth and TP 
degree: halving the pipeline depth reduces the warmup and 
cooldown bubble ratio, while the intra-node TP communication 
introduced by $D_t=2$ remains within acceptable bounds. (2) 
The $(4, 4)$ configuration attains the lowest throughput. 
The reason is that the larger TP degree pushes part of the 
TP communication across nodes, where the inter-node bandwidth 
becomes the dominant bottleneck. (3) Peak memory decreases 
substantially as $D_t$ grows, with doubling $D_t$ roughly 
halving the per-device peak each time and reducing the GPT 
peak from 60.1 GiB under $(16, 1)$ to 18.4 GiB under $(4, 4)$. 
The reason is that TP shards both 
parameter and activation memory across $D_t$ devices, 
reducing the per-device share by a factor of $D_t$ and 
effectively neutralizing BSIPipe's bidirectional memory 
overhead.

\subsection{Ablation Study}

To assess the contribution of each communication optimization 
in BSIPipe, we conduct ablation studies on two modified 
variants: one without the topology-aware device mapping and another 
without the forward-communication and backward-computation 
overlap. The deadlock avoidance with early receive is 
excluded, as removing it leaves the training in indefinite 
deadlock and unable to proceed.

Table~\ref{tab-ablation} presents the throughput results on 
16 GPUs at a sequence length of 20K with 64 micro-batches. 
Key observations include: (1) BSIPipe w/o device mapping 
incurs the larger throughput drop (1.59\% on GPT and 0.92\% 
on LLaMA3), underscoring the role of confining parameter 
gradient synchronization to high-bandwidth intra-node 
interconnects. (2) BSIPipe w/o F-B overlap exhibits a drop 
of 0.37\% on GPT and 0.60\% on LLaMA3, confirming the 
positive contribution of the forward-communication and 
backward-computation overlap to training throughput.

\begin{table}[tb]
\centering
\footnotesize
\setlength{\tabcolsep}{4pt}

\begin{tabular}{lcccc}
\toprule
\multirow{2}{*}{\textbf{Setting}} 
& \multicolumn{2}{c}{\textbf{Throughput}} 
& \multicolumn{2}{c}{\textbf{Drop vs.\ baseline}} \\
\cmidrule(lr){2-3} \cmidrule(lr){4-5}
& GPT & LLaMA3 & GPT & LLaMA3 \\
\midrule
BSIPipe (baseline)            & 68,954 & 73,453 & ---    & ---    \\
w/o device mapping            & 67,861 & 72,780 & 1.59\% & 0.92\% \\
w/o F-B overlap               & 68,702 & 73,015 & 0.37\% & 0.60\% \\
\bottomrule
\end{tabular}
\caption{Ablation of communication optimizations on 16 GPUs 
at a sequence length of 20K. Throughput in tokens/s. F-B 
overlap: forward-communication and backward-computation 
overlap.}
\label{tab-ablation}
\end{table}

\section{Conclusions}

In this paper, we propose BSIPipe, a bidirectional 
sequence-level interleaved pipeline parallelism for 
long-sequence LLM training. Extensive experiments on GPT and 
LLaMA3 across up to 32 GPUs show that BSIPipe achieves SOTA 
throughput with balanced memory distributions, demonstrating 
its scalability for large-scale distributed training. 
Furthermore, the comparison of parallel configurations shows 
that a suitable TP degree neutralizes BSIPipe's bidirectional 
memory overhead while maintaining throughput. 
Ablation studies confirm that the communication 
optimizations contribute positively to BSIPipe's 
performance.

\newpage

\bibliography{aaai2026nyz}

\end{document}